# Hashtag Healthcare: From Tweets to Mental Health Journals Using Deep Transfer Learning

Benjamin Shickel[1,a], Martin Heesacker[2,b], Sherry Benton[3,c], and Parisa Rashidi[4,d]

[1]Department of Computer and Information Science and Engineering, University of Florida, Gainesville, FL, 32611, USA
[2]Department of Psychology, University of Florida, Gainesville, FL, 32611, USA
[3]TAO Connect, Inc., St. Petersburg, FL, 33701, USA
[4]Department of Biomedical Engineering, University of Florida, Gainesville, FL, 32611, USA
[a]shickelb@ufl.edu
[b]heesack@ufl.edu
[c]sherry.benton@taoconnect.org
[d]parisa.rashidi@ufl.edu

**ABSTRACT**

As the popularity of social media platforms continues to rise, an ever-increasing amount of human communication and self-expression takes place online. Most recent research has focused on mining social media for public user opinion about external entities such as product reviews or sentiment towards political news. However, less attention has been paid to analyzing users' internalized thoughts and emotions from a mental health perspective. In this paper, we quantify the semantic difference between public Tweets and private mental health journals used in online cognitive behavioral therapy. We will use deep transfer learning techniques for analyzing the semantic gap between the two domains. We show that for the task of emotional valence prediction, social media can be successfully harnessed to create more accurate, robust, and personalized mental health models. Our results suggest that the semantic gap between public and private self-expression is small, and that utilizing the abundance of available social media is one way to overcome the small sample sizes of mental health data, which are commonly limited by availability and privacy concerns.

## Introduction

Sentiment analysis, the task referring to the automatic determination of user opinion from text, has received increased attention in the past decade[1–5]. Much of the success of sentiment analysis techniques can be attributed to the rise of social media platforms, where millions of users share their opinions on a wide variety of subjects. The majority of sentiment analysis methods are aimed at aggregating opinions towards entities like movies, people, products, or companies. We refer to this well-known research area as external sentiment analysis, in which sentiment and textual polarity is calculated with respect to a specific external entity. In contrast, we define internal sentiment analysis as the study of the polarity of user text with respect to themselves, primarily concerned with statements of emotion and mental health[6]. In this paper, we deal strictly with internal sentiment analysis, specifically with the valence prediction of private journals in a mental health therapy setting. Our work partly aligns with previous research regarding emotion detection in text[7–12], a subtask of the field of affective computing and analysis, but unlike previous work, we focus on the expansion of valence categories in a mental health setting.

One useful application domain of automated internal sentiment analysis frameworks is the burgeoning field of online mental health therapy services[13–23]. These programs provide the patient education components inherent in cognitive behavioral therapy (CBT). Practice identifying and changing unhelpful thought patterns or cognitive distortions is a central part of CBT. Ongoing practice with feedback can increase patients' ability to accurately identify more helpful and less helpful thoughts, but up until now, ongoing feedback has been difficult to provide. As part of online treatment, users typically submit several directed journals for documenting their daily thoughts and feelings as they improve their mental well-being using self-directed strategies taught as part of therapy. An accurate and validated system for automatically categorizing user text has obvious benefits to such a therapy service, such as flagging text which may be an early warning for suicide risk, providing a positive and always-available feedback for patients with distorted thinking, or simply providing enhanced and more fine-grained analysis of overall patient well-being.

Traditional sentiment analysis involves detecting whether a given text fragment is subjective or objective, and in the case of subjectivity, classifies the text as either positive or negative. While certainly the most straightforward design for external opinion mining, we take this analysis a step further for mental health polarity. Rather than framing the subjectivity identification task as a binary classification between positivity and negativity, we introduce two additional classes of polarity: **both** positive and negative, and **neither** positive nor negative. We made this decision based on psychological research, which suggests emotions cannot be represented on a single axis of valence[24–28]. Thus, text classified as neutral using traditional frameworks would, using our new annotation scheme, fall into either of the two augmented classes.

Unfortunately, publicly-available mental health datasets suitable for machine learning-based internal sentiment analysis are few and far between. However, large amounts of social media text have become available in recent years, and several studies have examined traditional sentiment analysis in the context of social media platforms such as Twitter[?, 29–34]. Given social media users' tendency towards self-expression, we hypothesize that the social media domain is quite similar to the mental health domain with regards to textual language modeling and classification, and can be used to help train mental health models and systems.

We quantify the similarity between the two domains using a machine learning technique known as transfer learning. This approach involves training models on one domain (i.e. social media text) and fine-tuning them on another target domain (i.e. mental health text). We specifically focus on transfer learning in the context of deep learning, a subfield of machine learning that is designed to automatically learn ideal representations of raw data without human supervision or manual feature engineering. In recent years, deep learning techniques have yielded state-of-the-art performance in the computer vision and natural language processing (NLP) domains. Success in NLP applications is due largely in part to these nonlinear models' ability to model language from raw characters or words.

We train deep learning models on a large dataset of annotated text from the social media platform Twitter, and transfer the underlying language model and learned representation to the task of predicting the valence of mental health journal text. Additionally, we use the same Twitter-based model to predict objective change in mental health as gauged by evidence-based mental health measures. We demonstrate that language models built from social media can be successfully transferred to the mental health domain for improved performance for emotional valence prediction, and argue that social media is a valuable source of data for text-based mental health applications, a domain in which labeled data is notoriously scarce.

## Methods

In this section, we provide implementation details, data description, and experimentation setup and methodology. We begin by describing the process by which we transfer knowledge from the social media domain to the mental health domain.

### Transfer learning

Generally, for machine learning applications when labeled data in the target domain is abundant, single-domain models are preferable due to their ability to recognize and generalize domain-specific patterns in the input. If labeled data is scarce, however, models have the tendency to overfit the training data by incorrectly attributing significance to small input variations and noise, and fail to generalize well. This is especially true for more complex models, including most deep learning techniques.

The primary idea behind transfer learning involves training a complex model on a *source* domain, which typically contains a vast amount of labeled data, and then *transferring* some or all of what the model has learned to a *target* domain. Much recent work on transfer learning has focused on deep learning techniques, which have proven highly successful in domains such as image and natural language processing. While the transfer learning process is possible with linear models like logistic regression or support vector machines, its benefits become more realized with deep learning methods, which inherently learn a characteristic domain representation from raw data. If the source and target domains are similar, this learned representation - which is expected to be more robust in the model trained on more data - can be utilized for similar tasks in other domains with data scarcity. For image processing, this representation could be the recognition of edges or shapes; for natural language processing, generalized notions of syntax and semantics can be transferred.

Our target domain includes individuals' text-based emotional self-expressions and reflections from a cognitive-behavioral perspective, obtained through an online mental health therapy service. Unfortunately, large public datasets relating to this field simply do not exist, which is our prime motivation for the application of transfer learning. We hypothesize that the social media domain, specifically the large amount of public tweets from Twitter, would be similar enough to transfer knowledge of content, style, and structure to the mental health domain.

Given recent advances in deep learning for a variety of natural language processing tasks, we explore the application of recurrent neural networks for transferring social media knowledge to the mental health domain. While a full review of deep learning is beyond the scope of this paper, we provide a general overview of our particular model in the following section.

### Recurrent neural networks

The particular algorithm we use to apply transfer learning techniques is the recurrent neural network (RNN), a class of deep learning architectures especially suited for processing sequential data. The RNN is a type of feedforward neural network that incorporates the notion of temporal or sequential memory, where the output from a single input time step is based not only on the input data itself, but also on the hidden representation from the previous time step. As a full input sequence is passed through the network, the final output is a repeated combination of the inputs and hidden representations from all input time steps.

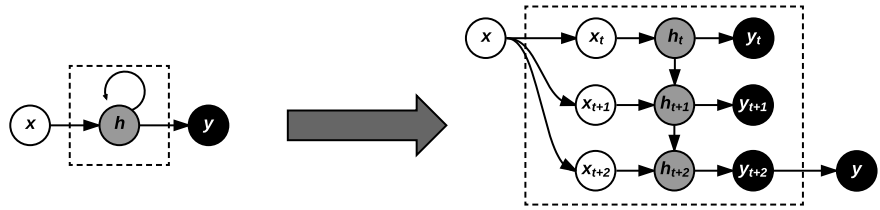

**Figure 1.** Symbolic (left) and expanded (right) representations of a traditional recurrent neural network (RNN) for sequence classification. Each time step of the input $x_i$ (in our paper, these are words) produces an output $y_i$, where the overall sequence prediction $y$ is obtained once the full sequence has passed through the network.

Equation 1 shows the calculation of an intermediate output $a$ at an arbitrary time step $t$ given an input $x$, where $W$ are learnable connection weights between hidden nodes, $U$ are learnable connection weights between inputs and hidden nodes, and $b$ is a bias term.

$$z^t = Wh^{t-1} + Ux^t + b \qquad (1)$$

The intermediate output $z^t$ for a single time step is then passed through a nonlinear function, such as tanh, to produce the RNN's activation for the given time step. For RNNs operating as supervised classifiers like ours, typically RNN activation of the final time step is fed to a fully-connected output layer of size $C$, where $C$ is the number of classes, upon which a softmax activation generates a prediction probability distribution over the available classes.

While we have briefly described the functioning of a traditional RNN, in this paper we use a model variant that uses gated recurrent units (GRU), a system of gates that controls information flow within the network, which has been shown to improve RNN performance by capturing longer-term dependencies and preventing vanishing gradients. Additionally, we include a word-level attention mechanism based on the work of Yang, et al.[35] that learns to focus on meaningful input words, to provide a measure of natural interpretability to our final model evaluation. Briefly, when an input sequence of length $n$ is passed through the deep network, each word $w_i$ in the sequence gets assigned an attention value $\alpha_i$ where $\sum_{i=1}^{n} \alpha_i = 1$. After the network is trained, we can examine the corresponding attention weights for examples the network has not previously seen, as a technique to determine the most influential words for each classification label.

### Datasets and annotation

Our source domain comes from the Sentiment140 Twitter dataset (http://www.sentiment140.com) designed for traditional sentiment analysis tasks. The data contains 1.6 million tweets annotated as positive or negative by an automatic process using the presence of specific emoticons. Our preprocessing steps included lowercasing, hyperlink removal, user mention removal, and punctuation stripping.

Our target domain includes IRB-approved responses retrospectively collected from monitoring logs submitted by users as part of the TAO Connect, Inc. online therapy assistance program (http://www.taoconnect.org). In total, we collected 3,872 total responses. We then employed three trained psychology graduates, under the supervision of departmental faculty, to provide valence labels for these responses. The annotators were given the option to annotate each response as positive, negative, both positive and negative, or neither positive nor negative. Final labels were assigned via a majority voting scheme, and in the case of a three-way tie, the response was discarded. The inter-rater agreement reliability (Cohen's kappa) between annotators 1 and 2 was 0.5, between annotators 2 and 3 was 0.67, and between annotators 1 and 3 was 0.48. The overall agreement reliability between all annotators (Fleiss' kappa) was 0.55.

Of the 3,872 responses, 63.5% were assigned the negative label (10.4% positive, 14.3% neither positive nor negative, 11.7% both positive and negative). The high class skew is a side effect of selection bias in our dataset - most psychotherapy seekers exhibit negative thinking, which is one major reason people seek therapy. We preprocessed the target domain text using the same methods as the source domain. The distribution of word counts for each of the two datasets is shown in Figure 3, where it is clear that respondents are mostly generating short text.

One example response from each of the four valence classes in our dataset is shown below. The examples illustrate subtle language cues that differentiate the classification of mental health text from traditional sentiment analysis tasks.

- **Positive:** *I am interesting and likable enough.*

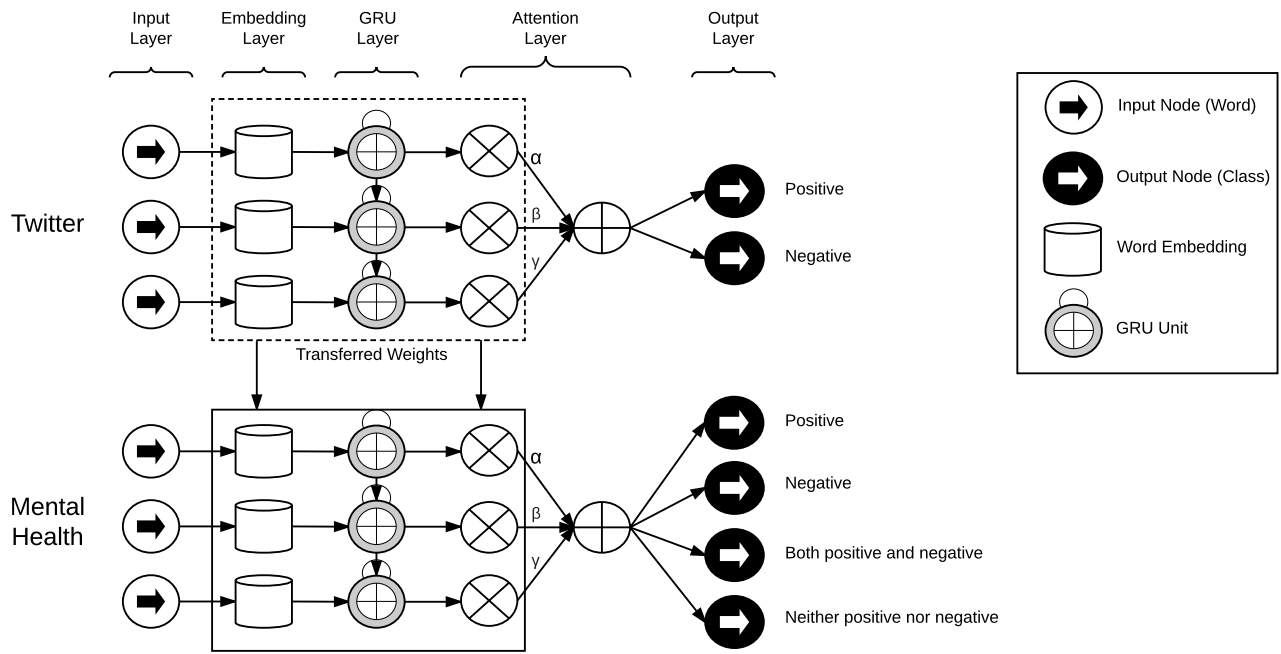

**Figure 2.** Framework for transferring knowledge from the social media domain to the mental health domain. After the binary classification model is trained solely on social media, we experiment with two methods of transferring knowledge to the new task with different classification objective: (1) initializing a new model's embedding layer with learned embeddings from Twitter, and (2) initializing both the embedding layer and GRU layers with attention with the corresponding weights from the source domain.

- **Negative:** *I don't do well at parties, I'm not interesting.*

- **Both positive and negative:** *I chewed my fingernails and did some skin picking. I tried doing some of the breathing exercises which worked while I did them.*

- **Neither positive nor negative:** *I wrote in my journal, and read till I was tired enough to fall asleep.*

## Experiments

Our experiments are designed to evaluate the performance gained by transferring social media knowledge to the mental health domain for the task of 4-way emotional valence prediction. In this paper, we compare two methods of language-based transfer learning involving social media data to two baseline approaches that operate solely in the target domain. The models included in our comparison are summarized below.

- **Single-domain linear classifier:** For this baseline model, we use the traditional NLP technique of vectorizing the mental health text using a bag-of-words approach, normalizing by term frequency-inverse document frequency (tf-idf). A logistic regression classifier is used to predict emotional valence from the vectorized text.

- **Single-domain RNN:** The second baseline model is a recurrent neural network (RNN) with gated recurrent units (GRU). All network weights are initialized randomly and learned only from the target domain as training progresses.

- **RNN with transferred word embeddings:** This model is the same as the single-domain neural network with one exception: we initialize the word embeddings with GloVe vectors pre-trained on Twitter text. This represents a more cautious approach to transferring knowledge, in which only semantic relationships between words are transferred, and not any knowledge about the sentiment of language derived from dataset-specific word sequences.

- **RNN with full weight transfer:** Our final model represents a full transfer of knowledge learned from social media. We first train an RNN on the Twitter dataset to classify tweets as positive or negative. We then remove the final layer from this model (a single output node with sigmoid activation and binary cross-entropy loss, corresponding to binary

4/10

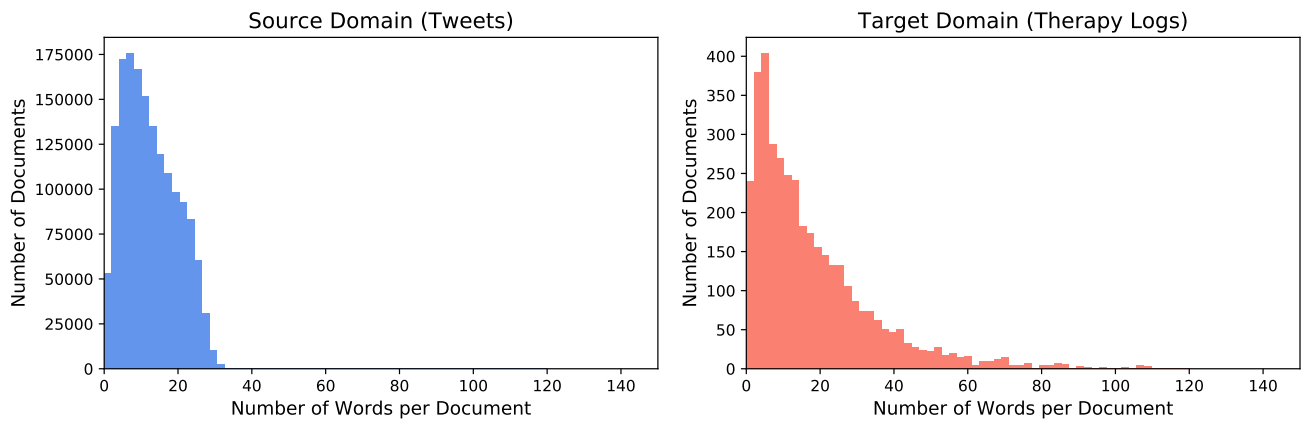

**Figure 3.** Word count distribution for both the source and target domain. Both domains consist of relatively short pieces of text. The longer tail of the mental health logs is due to the lack of character limits, unlike the Twitter dataset.

classification) and replace it with a four-node output layer with softmax activation and categorical cross-entropy loss, corresponding to the target domain's four-class classification task. In this manner, we re-use both the word embeddings and recurrent weights learned from the source domain.

For all models described above, we report 5-fold cross-validation results on the mental health dataset. Each fold is stratified to reflect the global distribution of valence classes. We performed a nested cross-validation procedure to simultaneously select optimal hyperparameters and evaluate the models, but found similar performance across a range of parameter values. For consistency, we set all RNN parameters equal to the optimal parameters from the Twitter model selected via 5-fold cross-validation. Namely, we use 50% dropout on word vectors, 64 bidirectional GRU units, tanh activation, 50% dropout on the RNN output, and use an Adam optimizer with learning rate of 0.01.

The datasets generated during and/or analyzed during the current study are not publicly available due to patient privacy concerns.

## Results

Given the large class imbalance of the mental health dataset, we chose macro F1 score as the primary metric for assessing model performance, and used this metric as our early stopping indicator for reducing overfitting. The RNN initialized with both word embeddings and recurrent weights from the binary Twitter dataset resulted in a macro F1 score of 0.68 on the four-class mental health dataset, an 11.5% improvement over the single-domain linear baseline. Additionally, this model scored best across all total metrics except for three, where the logistic regression classifier slightly outperformed in *Negative* recall, *Both Positive and Negative* precision, and *Neither Positive nor Negative* precision. Full results are shown in Table 1.

Overall, the performance of the RNN with transferred word embeddings fell in between the linear baseline and the full-weight transfer RNN on most metrics. The RNN without any transferred initializations performed poorly, and was outperformed by the linear baseline in several metrics. We show the area under the receiving operator characteristic curve (AUROC) for each valence category in Figure 4.

## Discussion

Both deep learning models that take advantage of learning from social media exhibit substantial increases over the linear baseline in recall for the extra valence classes (both positive and negative, neither positive nor negative), suggesting that the subtle cues differentiating these classes from the traditional "neutral" class are missed by simple linear models, and that learning traits of social media language use can be useful for mental health tasks.

While performing best in all aggregate metrics, the model with complete model transfer showed especially large improvements in AUC for the positive and negative classes (Figure 4), which demonstrates the semantic similarity of the target tasks of the both the source domain (external sentiment) and target domain (internal positivity and negativity).

The poor performance of the RNN without any knowledge transfer can be explained by the small size of the dataset, which resulted in large amounts of overfitting. Initializing the weights of the neural network with word embeddings improved the performance, and adding in recurrent weights from the transferred RNN model improved it even further.

| Class | Metric | Logistic Regression | RNN | RNN (Embedding Transfer) | RNN (Full Weight Transfer) |
|---|---|---|---|---|---|
| All | Macro F1 | 0.61 | 0.60 | 0.63 | **0.68** |
|  | Weighted F1 | 0.74 | 0.72 | 0.74 | **0.79** |
|  | Accuracy | 0.75 | 0.72 | 0.72 | **0.78** |
| Positive | Precision | 0.58 | 0.46 | 0.50 | **0.61** |
|  | Recall | 0.44 | 0.53 | 0.50 | **0.55** |
|  | F1 | 0.50 | 0.49 | 0.50 | **0.58** |
|  | AUC | 0.85 | 0.82 | 0.84 | **0.91** |
| Negative | Precision | 0.82 | 0.86 | 0.90 | **0.92** |
|  | Recall | **0.91** | 0.81 | 0.78 | 0.84 |
|  | F1 | 0.86 | 0.83 | 0.84 | **0.88** |
|  | AUC | 0.88 | 0.86 | 0.89 | **0.93** |
| Both | Precision | **0.50** | 0.42 | 0.39 | 0.47 |
|  | Recall | 0.39 | 0.48 | 0.61 | **0.62** |
|  | F1 | 0.44 | 0.44 | 0.47 | **0.54** |
|  | AUC | 0.85 | 0.78 | 0.85 | **0.87** |
| Neither | Precision | **0.69** | 0.63 | 0.67 | 0.66 |
|  | Recall | 0.59 | 0.67 | 0.73 | **0.79** |
|  | F1 | 0.64 | 0.65 | 0.70 | **0.72** |
|  | AUC | 0.91 | 0.91 | 0.93 | **0.94** |

**Table 1.** Results for both experimental tasks. The model with transferred social media knowledge achieves the best F1 scores and accuracy, as well as most of the class-specific evaluation metrics. As more information is transferred from the social media domain, the RNN-based models yield increasingly better performance, suggesting a semantic similarity between domains. Bold values correspond to the best-performing model for each metric.

In Table 2, we examine the RNN attention mechanism weights for words occurring in test set responses, and also show the distinction between instances based on the classifier's predicted valence class. The first and third columns of Table 2 clearly demonstrate the informativeness and weight of negative words relating to worry and anxiety for our mental health dataset, which is an expected outcome given the nature of mental health therapy services.

This study has important implications for computer-assisted, computerized psychotherapy, and other therapy-related applications in which patients submit narrative text. This tool can be used to identify circumstances where negative affect in patient narratives is high, suggesting elevated levels of distress, depression, or anxiety. It can also be used to detect cases in which positive and negative affect are both high in narrative text, which may be indicators of ambivalence, which is often associated with conflict-related distress. In addition to detecting static high levels of negative affect or high levels of both negative and positive affect, this tool can be used to detect dynamic changes across time. Declines in negative affect or declines in simultaneously high levels of positive and negative affect in narrative responses is an indication of symptom reduction and perhaps a sign of treatment effectiveness. In contrast, increases in negative affect or increases in simultaneously high levels of positive and negative affect in narrative responses is an indication of worsening symptoms and perhaps a sign of additional trauma and/or treatment ineffectiveness. These signs of worsening symptoms can signal the need to review and perhaps change treatment, to deal with recent additional trauma and/or to deal with adverse responses to specific interventions. Unchanging or minimally changing elevated levels of negative affect in written narratives across time indicate a lack of treatment effectiveness or a plateau in treatment, either of which may require making different treatment decisions. Cyclical oscillations in levels of negative affect or oscillations between high negative and high positive affect across time suggest the possibility of bi-polar disorder, cyclothymia, borderline personality disorder, and/or an interpersonal or social context fraught with cyclical affective instability. Thus, this tool can be useful in evaluating treatment effectiveness, evaluating when to change treatment strategies, detecting patient social stressors, and validating or invalidating initial clinical diagnoses. In addition, the tool can be used as an indication of treatment completion. When negative affect in narratives declines and remains low, clinicians should explore whether it is time to terminate therapy.

Not only can this tool be useful in making the general conclusions just discussed, but it can also be used to provide immediate feedback to patients, whether automated feedback or feedback from the therapist. Comments that reflect empathy for the patient's current affective state can produced by analyzing for dynamic changes or stability in the affect reflected in narrative responses. Patients whose affect is predominantly negative across time can be encouraged to hold on and prompted to accept that change often takes more time than one would like. Patients whose negative affect is declining or whose positive affect is increasing, or ideally both, can be prompted to savor and appreciate the improvement, and to reflect on the factors associated with that improvement. Those with cyclical patterns can be encouraged to be aware of and reflect on the oscillation and perhaps to explore factors contributing to the oscillation, and ways to cope with it.

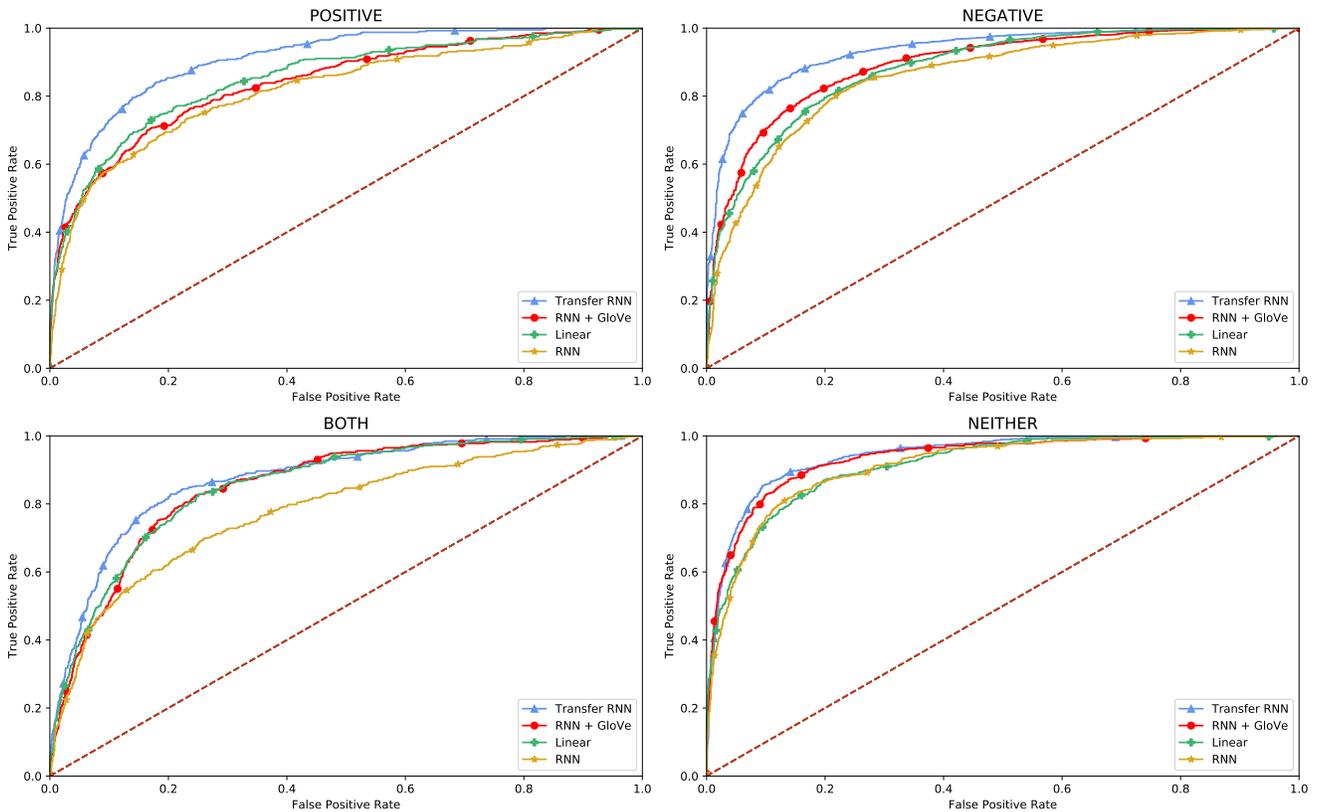

**Figure 4.** Area under the receiver operating characteristic curve (AUROC) for each valence class. The models with transferred social media knowledge perform best across all classes. The increase is most apparent for the *Positive* and *Negative* classes, which is unsurprising given that these were the two classes used to train the initial Twitter model.

The tool can also be used to help draw patients' attention to the affective conditions reflected in their narratives and to the patterns reflected in these narratives across time. This attention-drawing process can help patients access aspects of themselves that might otherwise be outside of their awareness. For example, patients who have struggled with long-term mental health concerns may not consciously realize they are improving until they receive feedback about the improving affective patterns reflected in the tool's analysis of their narratives. Another example is a patient who struggles with regular affective oscillations, but is unaware of the pattern across time and therefore limited in the ability to analyze, prevent, plan for, and cope with these cycles. A third example is that without feedback from the affect evaluation tool, patients and their therapists may not be aware of their deteriorating affective state until their situation is dangerous or catastrophic. A final example of patient self-awareness that can be enhanced by this tool involves premature termination. When patients still experience acute distress or significant clinical symptoms remain, receiving feedback that their narrative feedback is showing signs of improvement may encourage them to remain in treatment when they might otherwise drop out.

Yet another benefit of using this tool is the believability of the feedback. In situations in which patients might be skeptical of the accuracy of therapist feedback or reactive to the fact that a healthcare provider is providing feedback, those same patients might place their trust in the more objective feedback from the machine learning tool. In fact, there is evidence that ostensibly objective personality assessment feedback is widely accepted, especially if it is worded somewhat positively and somewhat generally[36] though future studies will need to be conducted to assess whether feedback from this tool is widely accepted by patients.

The tool can also be used by therapists and therapy developers to assess the effectiveness of specific computerized modules or specific interventions. And it can be used to assess whether these modules and interventions are differentially effective for different types of patients, for different types of presenting concerns, at different times in the therapy process, and whether different sequences of modules or interventions produce different affective responses in patient narratives. Finally, the tool can be used to assess whether particular identifiable affective patterns emerge over time from patients diagnosed with different disorders.

| All Classes | | Positive | | Negative | | Both | | Neither | |
|---|---|---|---|---|---|---|---|---|---|
| worried | 446.1 | strangers | 17.5 | worried | 441.4 | better | 137.8 | but | 125.7 |
| not | 432.3 | didnt | 16.4 | not | 389.3 | fine | 56.0 | down | 38.2 |
| anxious | 310.9 | not | 14.7 | anxious | 302.0 | calm | 52.7 | tried | 31.1 |
| didnt | 262.5 | situation | 13.3 | uncomfortable | 254.7 | likeable | 50.8 | bad | 30.7 |
| uncomfortable | 255.5 | have | 9.1 | didnt | 237.6 | great | 43.1 | not | 21.6 |
| bad | 221.7 | i | 9.1 | bad | 190.6 | enough | 31.0 | better | 18.0 |
| but | 179.3 | the | 9.1 | awkward | 134.4 | assure | 27.3 | i | 17.9 |
| stressed | 135.9 | was | 9.0 | stressed | 132.3 | interesting | 25.6 | okay | 16.9 |
| awkward | 135.3 | strife | 8.3 | couldnt | 107.2 | well | 24.9 | calm | 16.2 |
| couldnt | 109.9 | busy | 6.6 | nervous | 105.7 | relaxed | 23.3 | and | 15.3 |
| nervous | 107.7 | thoughts | 6.1 | wont | 98.5 | worries | 22.5 | a | 14.3 |
| overwhelmed | 83.5 | a | 5.9 | cant | 83.4 | and | 19.5 | worry | 14.1 |
| anxiety | 76.1 | work | 5.7 | overwhelmed | 81.8 | nice | 16.1 | have | 13.7 |

**Table 2.** Largest sums of attention weights (x $10^4$) for each word appearing in samples from the unseen mental health testing set, using the best-performing model with full weight transfer from the social media to mental health domain. In each column, the sums are normalized by the global sum of all generated attention values. The first column contains the words with largest attention values across the entire testing set. The remaining columns group the words by the valence class predicted by the model. Words relating to worry, anxiety, and overall negativity garner the highest attention values, and their presence serves as an immediate indicator of negative valence.

Through a technical comparison based on deep learning, we have shown that public social media posts can be a viable, if not necessary, inclusion for more robust mental health models. Given the overall performance increases after applying transfer learning, it is reasonable to assume that semantics between public and private self-expression are similar, especially in the context of emotional valence. However, our study is not without limitations. While our experiments help shed light on the minimal semantic gap between social media and mental health, we are only focusing on the particular task of emotional valence prediction. Additionally, our target domain data came from a single institution, which resulted in a particular class skew that might not reflect global populations. In future work, we plan to explore a variety of mental health tasks and applications such as suicide prevention and identification of distorted thought patterns. Additionally, we plan to explore the inverted transfer setting, where private mental health journals may prove useful for predictive tasks in the public social media domain.

## Acknowledgements

We thank TAO Connect, Inc. for access and assistance with retrieving only therapy logs. The Titan X Pascal partially used for this research was donated by the NVIDIA Corporation.